# Large-Sample Learning of Bayesian Networks is NP-Hard

David Maxwell Chickering
Microsoft Research
Redmond, WA 98052
dmax@microsoft.com

Christopher Meek
Microsoft Research
Redmond, WA 98052
meek@microsoft.com

David Heckerman
Microsoft Research
Redmond, WA 98052
heckerma@microsoft.com

**Abstract**

In this paper, we provide new complexity results for algorithms that learn discrete-variable Bayesian networks from data. Our results apply whenever the learning algorithm uses a scoring criterion that favors the simplest model able to represent the generative distribution exactly. Our results therefore hold whenever the learning algorithm uses a consistent scoring criterion and is applied to a sufficiently large dataset. We show that identifying high-scoring structures is NP-hard, even when we are given an independence oracle, an inference oracle, and/or an information oracle. Our negative results also apply when learning discrete-variable Bayesian networks in which each node has at most $k$ parents, for all $k \geq 3$.

## 1 INTRODUCTION

Researchers in the UAI community have generally accepted that without restrictive assumptions, learning Bayesian networks from data is NP-hard, and consequently a large amount of work in this community has been dedicated to heuristic-search techniques to identify good models. A number of discouraging complexity results have emerged over the last few years that indicate that this belief is well founded. Chickering (1996) shows that for a general (and widely used) class of Bayesian scoring criteria, identifying the highest-scoring structure from small-sample data is hard, even when each node has at most two parents. Dasgupta (1999) shows that it is hard to find the polytree with highest maximum-likelihood score. Although we can identify the highest-scoring tree structure using a polynomial number of calls to the scoring criterion, Meek (2001) shows that identifying the best *path structure*—that is, a tree in which each node has degree at most two—is hard. Bouckaert (1994) shows that for independence-based learning algorithms, finding the simplest model that is consistent with an independence oracle is hard.

In this paper, we are interested in the large-sample version of the learning problem considered by Chickering (1996). The approach used by Chickering (1996) to reduce a known NP-complete problem to the problem of learning is to construct a complicated prior network that defines the Bayesian score, and then create a dataset consisting of a single record. Although the result is discouraging, the proof technique leaves open the hope that, in scenarios where the network scores are more "well behaved", learning is much easier.

As the number of records in the observed data grows large, most scoring criteria will agree on the same partial ranking of models; in particular, any *consistent* scoring criterion will—in the limit of large data—favor a model that can represent the generative distribution over a model that cannot, and when comparing two models that can both represent the generative distribution, will favor the model with fewer parameters. Almost all scoring criteria used in practice are consistent, including (1) any Bayesian criterion that does not rule out models apriori, (2) the minimum-description-length criterion, and (3) the Bayesian-information criterion.

In this paper, we consider the scenario when a learning algorithm is using a consistent scoring criterion with a large dataset. We assume that the learning algorithm has direct access to the generative distribution itself; the learning problem is thus to identify the simplest model that can represent that distribution exactly.

There are scenarios in which we can accomplish large-sample learning efficiently. If (1) there exists a DAG-model solution in which all independence and dependence relationships implied by that model hold in the generative distribution (that is, the generative distribution is *DAG perfect* with respect to the observable variables) and (2) we know that there exists such a solution in which each node has at most $k$ parents (for some constant $k$), then we can apply the SGS algorithm of Spirtes, Glymour and Scheines (2000) to identify the best network structure in a polynomial number of independence tests. In particular, because we know the value $k$, we can limit the worst-case number of independence tests used by the algorithm. Alternatively, if (1) the generative distribution is DAG perfect with respect to *some* DAG model (which might contain unobserved variables) and (2) we are given a total ordering over the variables that is consistent with the best structure, then we can find the best DAG model using a polynomial number of calls to the scoring criterion. In particular, we can apply a version of the GES algorithm of Meek (1997) that greedily adds and then deletes the parents of each node.

Unfortunately, the assumptions needed for these special-case efficient solutions are not likely to hold in most real-world scenarios. In this paper, we show that in general—without the assumption that the generative distribution is DAG perfect with respect to the observables and without the assumption that we are given a total ordering—large-sample learning is NP-hard. We demonstrate that learning is NP-hard even when (1) we are given an independence oracle, (2) we are given given an inference oracle, or (3) we are given an information oracle. We show that these results also apply to the problem of identifying high-scoring structures in which each node has at most $k$ parents, for all $k \geq 3$.

## 2 BACKGROUND

In this section, we provide background material relevant to the rest of the paper. We denote a variable by an upper case token (e.g., $A, B_i, Y$) and a state or value of that variable by the same token in lower case (e.g., $a, b_i, y$). We denote sets with bold-face capitalized tokens (e.g., $\mathbf{A}, \mathbf{B}$) and corresponding sets of values by bold-face lower case tokens (e.g., $\mathbf{a}, \mathbf{b}$). Finally, we use calligraphic tokens (e.g., $\mathcal{M}$) to denote statistical models and graphs.

In this paper, we concentrate on Bayesian networks for a set of variables $\mathbf{X} = \{X_1, \ldots, X_n\}$, where each $X_i \in \mathbf{X}$ has a finite number of states. A *parametric Bayesian-network model* for a set of variables $\mathbf{X}$ is a pair $(\mathcal{G}, \theta_\mathcal{G})$ that defines a joint probability distribution over $\mathbf{X}$. $\mathcal{G} = (\mathbf{V}, \mathbf{E})$ is a directed acyclic graph—or *DAG* for short—consisting of (1) nodes $\mathbf{V}$ in one-to-one correspondence with the variables $\mathbf{X}$, and (2) directed edges $\mathbf{E}$ that connect the nodes. $\theta_\mathcal{G}$ is a set of parameter values that specify the conditional probability distributions that collectively define the joint distribution.

We assume that each conditional probability distribution is a full table. That is, for each variable there is a separate (unconstrained) multinomial distribution given every distinct configuration of the parent values. For a variable $X_i$ with $r_i$ states, $r_i - 1$ parameters are both necessary and sufficient to specify an arbitrary multinomial distribution over $X_i$. Thus, assuming that there are $q_i$ distinct parent configurations for $X_i$, the conditional distribution for $X_i$ will contain $(r-1) \cdot q_i$ parameter values. We also assume that the number of states for each variable is some constant that does not depend on the number of variables in the domain.

A *DAG model* $\mathcal{G}$ is a directed acyclic graph and represents a family of distributions that satisfy the independence constraints that must hold in any distribution that can be represented by a parametric Bayesian network with that structure. We say that a DAG model $\mathcal{G}$ *includes* a distribution $p(\mathbf{X})$—and that $p(\mathbf{X})$ is *included by* $\mathcal{G}$—if the distribution can be represented by some parametric Bayesian network with structure $\mathcal{G}$.

Pearl (1988) provides a graphical condition called *d-separation* that can be used to identify any independence constraint implied by a DAG model; we assume that the reader is familiar with this condition. We use $X \perp\!\!\!\perp_\mathcal{G} Y | \mathbf{Z}$ to denote the assertion that DAG $\mathcal{G}$ imposes the constraint—via d-separation—that for all values $\mathbf{z}$ of the set $\mathbf{Z}$, $X$ is independent of $Y$ given $\mathbf{Z} = \mathbf{z}$. For a probability distribution $p(\cdot)$, we use $X \perp\!\!\!\perp_p Y | \mathbf{Z}$ to denote the assertion that for all values $\mathbf{z}$ of the set $\mathbf{Z}$, $X$ is independent of $Y$ given $\mathbf{Z} = \mathbf{z}$ in $p$.

We say that a distribution $p(\mathbf{X})$ is *perfect* with respect to a DAG model $\mathcal{G}$ if the both the independence and dependence relationships implied by d-separation in $\mathcal{G}$ hold in $p(\mathbf{X})$. We say that $p(\mathbf{X})$ is *DAG perfect* if there exists a DAG $\mathcal{G}$ such that $p(\mathbf{X})$ is perfect with respect to $\mathcal{G}$.

We say that two DAG models $\mathcal{G}$ and $\mathcal{G}'$ are *equivalent* if the two sets of distributions included by $\mathcal{G}$ and $\mathcal{G}'$ are the same. Because we are using complete tables as conditional distributions, an equivalent definition for the class of DAG models that we consider is that $\mathcal{G}$ and $\mathcal{G}'$ are equivalent if they impose the same independence constraints. For any DAG $\mathcal{G}$, we say an edge $X \rightarrow Y$ is *covered* in $\mathcal{G}$ if $X$ and $Y$ have identical parents, with the exception that $X$ is not a parent of itself. The significance of covered edges is evident from the following result:

**Lemma 1 (Chickering, 1995)** *Let $\mathcal{G}$ be any DAG model, and let $\mathcal{G}'$ be the result of reversing the edge $X \rightarrow Y$ in $\mathcal{G}$. Then $\mathcal{G}'$ is a DAG that is equivalent to $\mathcal{G}$ if and only if $X \rightarrow Y$ is covered in $\mathcal{G}$.*

As described above, when a parametric Bayesian network has complete tables, the number of parameters is completely determined by its DAG. Thus, we say that a DAG model *supports* a number of parameters $k$ when all parametric Bayesian networks with that structure contain $k$ parameters. The following result follows immediately from Lemma 1 for models with complete tables:

**Lemma 2 (Chickering, 1995)** *If $\mathcal{G}$ and $\mathcal{G}'$ are equivalent, then they support the same number of parameters.*

We say that a DAG model $\mathcal{H}$ includes a DAG model $\mathcal{G}$ if every distribution included in $\mathcal{G}$ is also included in $\mathcal{H}$. As above, an alternative but equivalent definition is that $\mathcal{H}$ includes $\mathcal{G}$ if every independence constraint implied by $\mathcal{H}$ is also implied by $\mathcal{G}$. Note that we are using "includes" to describe the relationship between a model and a particular distribution, as well as a relationship between two models.

**Theorem 3 (Chickering, 2002)** *If $\mathcal{G}$ includes $\mathcal{F}$, then there exists a sequence of single edge additions and covered edge reversals in $\mathcal{F}$ such that (1) after each addition and reversal, $\mathcal{F}$ remains a DAG, (2) after each addition and reversal, $\mathcal{G}$ includes $\mathcal{F}$, and (3) after all additions and reversals, $\mathcal{F} = \mathcal{G}$.*

The "converse" of Theorem 3 will also prove useful.

**Lemma 4** *If $\mathcal{F}$ can be transformed into $\mathcal{G}$ by a series of single edge additions and covered edge reversals, such that after each addition and reversal $\mathcal{F}$ remains a DAG, then $\mathcal{G}$ includes $\mathcal{F}$.*

**Proof:** Follows immediately from Lemma 1 and from the fact that the DAG $\mathcal{F}'$ that results from adding a single edge to $\mathcal{F}$ necessarily includes $\mathcal{F}$. □

We now define the decision problems that we use to prove that learning is NP-hard. As discussed in Section 1, in the limit of large data, all consistent scoring criteria rank models that include the generative distribution over those that do not, and among those models that include the generative distribution, the criteria rank according to the number of parameters supported (with simpler models receiving better scores). Thus, a natural decision problem corresponding to large-sample learning is the following:

**LEARN**
INSTANCE: Set of variables $\mathbf{X} = \{X_1, \ldots, X_n\}$, probability distribution $p(\mathbf{X})$, and constant parameter bound $d$.
QUESTION: Does there exist a DAG model that includes $p$ and supports $\leq d$ parameters?

It is easy to see that if there exists an efficient algorithm for learning the optimal DAG model from large-sample data, we can use that algorithm to solve LEARN: simply learn the best model and evaluate the number of parameters it supports. By showing that LEARN is NP-hard, we therefore immediately conclude that the optimization problem of *identifying* the optimal DAG model is hard as well. We show that LEARN is NP-hard using a reduction from a restricted version of the NP-complete problem FEEDBACK ARC SET. The general FEEDBACK ARC SET problem is stated by Garey and Johnson (1979) as follows:

**FEEDBACK ARC SET**
INSTANCE: Directed graph $\mathcal{G} = (\mathbf{V}, \mathbf{A})$, positive integer $k \leq |\mathbf{A}|$.
QUESTION: Is there a subset $\mathbf{A}' \subset \mathbf{A}$ with $|\mathbf{A}'| \leq k$ such that $\mathbf{A}'$ contains at least one arc from every directed cycle in $\mathcal{G}$?

Garvill (1977) shows that FEEDBACK ARC SET remains NP-complete for directed graphs in which no vertex has a total in-degree and out-degree more than three. We refer to this restricted version as DEGREE-BOUNDED FEEDBACK ARC SET, or *DBFAS* for short.

## 3 MAIN RESULTS

We now provide the main results of this paper. In Section 3.1, we describe a polynomial-time reduction from instances of DBFAS to instances of LEARN. In Section 3.2, we prove that there is a solution to an instance of DBFAS if and only if there is a solution to the instance of LEARN that results from the reduction, and therefore we establish that LEARN is NP-hard. In Section 3.3, we extend our main result to the case when the learning algorithm has access to various oracles, and to the case when there is an upper bound on the number of parents for each node in the solution to LEARN.

For the remainder of this paper we assume—without loss of generality—that in any instance of DBFAS, no vertex has in-degree or out-degree of zero; if such a node exists, none of its incident edges can participate in a cycle, and we can remove that node from the graph without changing the solution.

### 3.1 A REDUCTION FROM DBFAS TO LEARN

In this section, we show how to reduce an arbitrary instance of DBFAS into a corresponding instance of LEARN. To help distinguish between elements in the instance of DBFAS and elements in the instance of LEARN, we will subscript the corresponding symbols with 'D' and 'L', respectively. In particular, we use $\mathcal{G}_D = (\mathbf{V}_D, \mathbf{A}_D)$ and $k_D$ to denote the graph and arc-set bound, respectively, from the instance of DBFAS; from this instance, we create an instance of LEARN consisting of a set of variables $\mathbf{X}_L$, a probability distribution $p_L(\mathbf{X}_L)$, and a parameter bound $d_L$.

For each $V_i \in \mathbf{V}_D$ in the instance of DBFAS, we create a corresponding nine-state discrete variable $V_i$ for $\mathbf{X}_L$. For each arc $V_i \to V_j \in \mathbf{A}_D$ in the instance of DBFAS, we create seven discrete variables for $\mathbf{X}_L$: $A_{ij}, B_{ij}, C_{ij}, D_{ij}, E_{ij}, F_{ij}, G_{ij}$. Variables $A_{ij}$, $D_{ij}$ and $G_{ij}$ have nine states, variables $B_{ij}$, $E_{ij}$ and $F_{ij}$ have two states, and variable $C_{ij}$ has three states. There are no other variables in $\mathbf{X}_L$ for the instance of LEARN. The probability distribution $p_L(\mathbf{X}_L)$ for the instance of LEARN is specified using a parametric Bayesian-network model $(\mathcal{H}_L, \theta_{\mathcal{H}_L})$. The model is defined over the variables in $\mathbf{X}_L$, along with, for each arc $V_i \to V_j \in \mathbf{A}_D$ from the instance of DBFAS, a single "hidden" binary variable $H_{ij}$. Let $\mathbf{H}_L$ denote the set of all such hidden variables. The distribution $p_L(\mathbf{X}_L)$ is defined by summing the distribution $p_L(\mathbf{H}_L, \mathbf{X}_L)$, defined by $(\mathcal{H}_L, \theta_{\mathcal{H}_L})$, over all of the variables in $\mathbf{H}_L$. The structure $\mathcal{H}_L$ is defined as follows. For each arc $V_i \to V_j \in \mathbf{A}_D$ in the instance of DBFAS, the DAG contains the edges shown in Figure 1. The number of states for each node in the figure is specified in parentheses below the node.

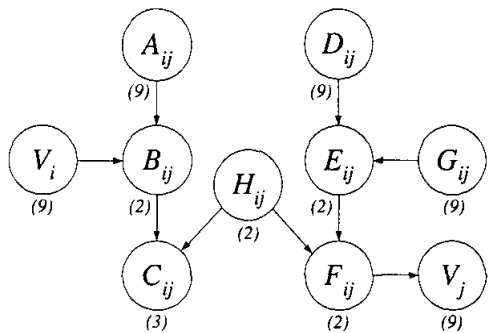

Figure 1: Edges in $\mathcal{H}_L$ corresponding to each arc $V_i \to V_j \in \mathbf{A}_D$ from the instance of DBFAS. The number of states for each node is given in parentheses below the node.

In Figure 2, we give an example of the structure of $\mathcal{H}_L$ that results from a specific instance of DBFAS.

We now specify the local probability distributions in $\theta_{\mathcal{H}_L}$. Let $r_X$ denote the number of states of $X$, let $\mathbf{Pa}_X$ denote the set of parents of $X$ in $\mathcal{H}_L$, and let $NZ(\mathbf{pa}_X)$ denote the number of values in $\mathbf{pa}_X$ that are equal to zero. Then for each node $X$ in $\mathcal{H}_L$, the local probability distribution for $X$ is defined as follows:

$$p(X = x | \mathbf{Pa}_X = \mathbf{pa}_X) \quad (1)$$
$$= \begin{cases} \frac{1}{16} & \text{if } x = 0 \text{ and } NZ(\mathbf{pa}_X) = 2 \\ \frac{1}{(r_X - 1)}\frac{15}{16} & \text{if } x \neq 0 \text{ and } NZ(\mathbf{pa}_X) = 2 \\ \frac{1}{64} & \text{if } x = 0 \text{ and } NZ(\mathbf{pa}_X) = 1 \\ \frac{1}{(r_X - 1)}\frac{63}{64} & \text{if } x \neq 0 \text{ and } NZ(\mathbf{pa}_X) = 1 \\ \frac{1}{128} & \text{if } x = 0 \text{ and } NZ(\mathbf{pa}_X) = 0 \\ \frac{1}{(r_X - 1)}\frac{127}{128} & \text{if } x \neq 0 \text{ and } NZ(\mathbf{pa}_X) = 0 \end{cases}$$

Because each node in $\mathcal{H}_L$ has at most two parents, the above conditions define every local distribution in $\theta_{\mathcal{H}_L}$.

Finally, we define the constant $d_L$ in the instance of LEARN. Every node in $\mathcal{G}_D$ has either exactly one or exactly two parents because, in any instance of DBFAS, the total degree of each node is at most three and by assumption no node has an in-degree or an out-degree of zero. Let $t_D$ denote the number of nodes in $\mathcal{G}_D$ from the instance of DBFAS that have exactly two in-coming edges; similarly, let $o_D = |\mathbf{V}_D| - t_D$ be the number of nodes that have exactly one in-coming edge. Then we have:

$$d_L = 186|\mathbf{A}_D| + 18k_D + 16(|\mathbf{A}_D| - k_D) + 16o_D + 32t_D \quad (2)$$

We now argue that the reduction is polynomial. It is easy to see that we can specify the structure of $\mathcal{H}_L$ and the bound $d_L$ in polynomial time; we now argue that we can specify all of the parameters $\theta_{\mathcal{H}_L}$ in polynomial time as well. Because each node in $\mathcal{H}_L$ has at most two parents, each corresponding conditional-probability table contains a constant number of parameters. Thus, as long as each parameter is represented using number of bits that is polynomial in the size of the instance of DBFAS, the parameters $\theta_{\mathcal{H}_L}$ can be written down in polynomial time. Each node has either two, three, or nine states, and thus it follows from the specification of $p(X = x | \mathbf{Pa}_X = \mathbf{pa}_X)$ in Equation 1 that each parameter is a fraction whose denominator is a power of two that can never exceed 1024 (i.e. $(9-1) \times 128$). Thus, when using a straightforward binary representation for the parameter values, we can represent each such value exactly using at most ten (i.e. $\log_2 1024$) bits. Thus we conclude that the entire reduction is polynomial.

## 3.2 REDUCTION PROOFS

In this section, we prove LEARN is NP-hard by demonstrating that there is a solution to the instance of DBFAS if and only if there is a solution to the instance of LEARN that results from the reduction. In the results that follow, we often consider sub-graphs of solutions to LEARN that correspond only to those nodes that are "relevant" to a particular arc in the instance of DBFAS. Therefore, to simplify the discussion, we use $\{V_i, V_j\}$ *edge component* to refer to a sub-graph defined by the nodes $\{V_i, A_{ij}, B_{ij}, C_{ij}, D_{ij}, E_{ij}, F_{ij}, G_{ij}, V_j\}$. We use *edge component* without reference to a particular $V_i$ and $V_j$ when an explicit reference is not necessary. Figure 3, which is key to the results that follow, shows two configurations of the edges in an edge component.

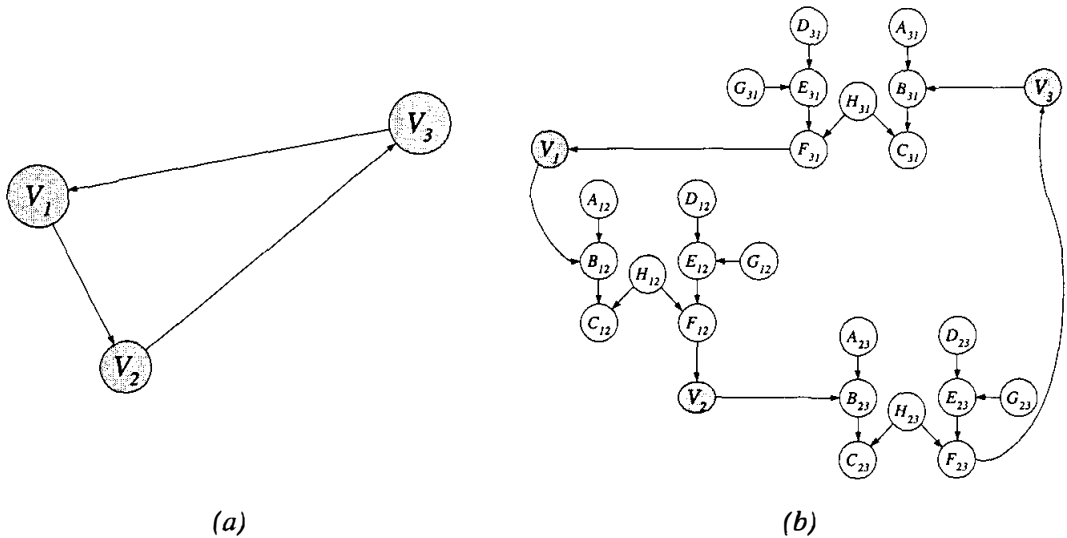

(a)            (b)

Figure 2: An example of the structure $\mathcal{H}_L$ that results from the reduction from a specific instance of DBFAS: (a) an instance of DBFAS consisting of three nodes $V_1$, $V_2$ and $V_3$ and (b) the corresponding structure of $\mathcal{H}_L$.

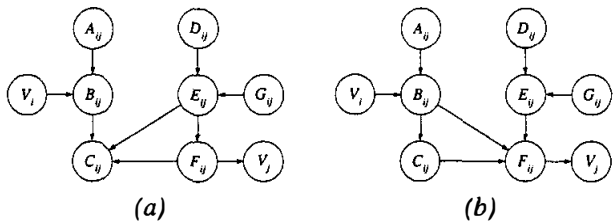

(a)            (b)

Figure 3: Two configurations of the edges in an edge component.

The first result, which is proved by Chickering, Meek and Heckerman (2003), demonstrates that the parameterization of $\mathcal{H}_L$ given in Equation 1 leads to a distribution $p_L(\mathbf{H}_L, \mathbf{X}_L)$ that is perfect with respect to $\mathcal{H}_L$.

**Theorem 5 (Chickering et al., 2003)**
*Let $(\mathcal{G}, \theta_\mathcal{G})$ be a parametric Bayesian-network model, let $r_X$ denote the number of states of node $X$, let $\mathbf{Pa}_X$ denote the set of parents of node $X$ in $\mathcal{G}$, let $N1(\mathbf{pa}_X)$ denote the number of non-zero elements in the set $\mathbf{pa}_X$, and let $\alpha$ be a constant satisfying $0 < \alpha < 1$. If all of the local distributions in $\theta_\mathcal{G}$ are defined as:*

$$p(X = x | \mathbf{Pa}_X = \mathbf{pa}_X) \qquad (3)$$
$$= \begin{cases} \alpha^{F(\mathbf{pa}_X)} & \text{if } x = 0 \\ \frac{1}{(r_X - 1)} \left(1 - \alpha^{F(\mathbf{pa}_X)}\right) & \text{otherwise} \end{cases}$$

*where*

$$F(\mathbf{pa}_X) = 2 - 2^{-N1(\mathbf{pa}_X)}$$

*then the distribution defined by $(\mathcal{G}, \theta_\mathcal{G})$ is perfect with respect to $\mathcal{G}$.*

To explicitly connect Theorem 5 to our proof, we provide the following simple extension:

**Corollary 6** *The distribution $p_L(\mathbf{H}_L, \mathbf{X}_L)$ resulting from the reduction is perfect with respect to $\mathcal{H}_L$.*

**Proof:** Follows because when $\alpha = 1/16$, Equation 3 results in identical distributions to Equation 1. □

The next result is used in both of the main proofs of this section. We assume the reader is familiar with the definition of an *active path* that defines the d-separation criterion. Recall that $\mathcal{H}_L$ contains an additional "hidden" node $H_{ij}$ within each edge component. We will be considering active paths in $\mathcal{H}_L$, but are only concerned about those in which the endpoints are in $\mathbf{X}_L$ and for which no $H_{ij}$ is in the conditioning set; these active paths correspond to dependencies that exist within the (marginalized) distribution $p_L(\mathbf{X}_L)$. To simplify presentation, we define a $\mathbf{X}_L$-*restricted* active path to denote such an active path. In this and later results, we will demonstrate that one DAG $\mathcal{F}_1$ includes another DAG $\mathcal{F}_2$ by showing that for any active path in $\mathcal{F}_2$, there exists a corresponding (i.e. same endpoints and same conditioning set) active path in $\mathcal{F}_1$.

**Lemma 7** *Let $p_L(\mathbf{X}_L)$ be the distribution defined for the instance of LEARN in the reduction, and let $\mathcal{F}$ be any DAG defined over $\mathbf{X}_L$ such that each edge component in $\mathcal{F}$ contains the edges in either Figure 3a or in Figure 3b. Then $\mathcal{F}$ includes $p_L(\mathbf{X}_L)$.*

**Proof:** Let $\mathcal{H}_L$ be the DAG defining $p_L(\mathbf{X}_L)$ in the reduction. We prove that $\mathcal{F}$ includes $p_L(\mathbf{X}_L)$ by demonstrating that for every $\mathbf{X}_L$-restricted active path in $\mathcal{H}_L$, there exists a corresponding active path in $\mathcal{F}$. To do this, we construct an additional model $\mathcal{H}'$ that includes $\mathcal{H}_L$—and consequently $\mathcal{H}'$ can represent $p_L(\mathbf{X}_L)$ exactly—such that $\mathbf{X}_L$-restricted active paths in $\mathcal{H}'$ are easily mapped to their corresponding active paths in $\mathcal{F}$.

We create $\mathcal{H}'$ as from $\mathcal{H}_L$ follows. For each $i$ and $j$, if the edge component in $\mathcal{F}$ is in the configuration shown in Figure 3a, we add the edge $E_{ij} \to H_{ij}$ to $\mathcal{H}$ and then reverse the (now covered) edge $H_{ij} \to F_{ij}$. Similarly, if the edge component in $\mathcal{F}$ is in the configuration shown in Figure 3b, we add the edge $B_{ij} \to H_{ij}$ to $\mathcal{H}$ and then reverse the edge $H_{ij} \to C_{ij}$. The resulting components in $\mathcal{H}'$ are shown in Figure 4a and Figure 4b, respectively. Because we created $\mathcal{H}'$ by edge additions and covered edge reversals, we know by Lemma 4 that $\mathcal{H}'$ includes $\mathcal{H}_L$. It is now easy to see that any $\mathbf{X}_L$-restricted active path in $\mathcal{H}'$ has a corresponding active path in $\mathcal{F}$: simply replace any segment $X \to H_{ij} \to Y$ in the path by the corresponding edge $X \to Y$ from $\mathcal{F}$, and the resulting path will be active in $\mathcal{F}$. □

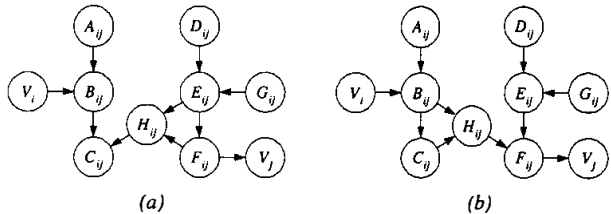

Figure 4: Edges in $\mathcal{H}'$ corresponding to the edge components in Figure 3

**Theorem 8** *There is a solution $\mathcal{F}_L$ to the instance of LEARN with $\leq d_L$ parameters if there is a solution $\mathbf{A}'_D$ to the given instance of DBFAS with $|\mathbf{A}'_D| \leq k_D$.*

**Proof:** We create a solution DAG $\mathcal{F}_L$ as follows. For every arc $V_i \to V_j \in \mathbf{A}'_D$ in the DBFAS solution, $\mathcal{F}_L$ contains the edges shown in Figure 3a. For the remaining arcs $V_i \to V_j$ that are not in $\mathbf{A}'_D$, $\mathcal{F}_L$ contains the edges shown in Figure 3b. $\mathcal{F}_L$ contains no other edges. First we argue that $\mathcal{F}_L$ is acyclic. Each $\{V_i, V_j\}$ edge component in $\mathcal{F}_L$ is itself acyclic, and contains a directed path from $V_i$ to $V_j$ if and only if the corresponding arc $V_i \to V_j \in \mathbf{A}_D$ from the instance of DBFAS is not in $\mathbf{A}'_D$; if the corresponding arc from the instance of DBFAS is in $\mathbf{A}'_D$, $\mathcal{F}_L$ contains neither a directed path from $V_i$ to $V_j$, nor a directed path from $V_j$ to $V_i$ that is contained within the edge component. Therefore, for any hypothetical cycle in $\mathcal{F}_L$, there would be a corresponding cycle in $\mathcal{G}_D$ that passed entirely through arcs not in $\mathbf{A}'_D$, which is impossible assuming $\mathbf{A}'_D$ is a solution to DBFAS. From Lemma 7, we know that $\mathcal{F}_L$ includes $p_L(\mathbf{X}_L)$. Now we derive the number of parameters supported by $\mathcal{F}_L$. Within each edge component, the parents for $A_{ij}$, $B_{ij}$, $D_{ij}$, $E_{ij}$ and $G_{ij}$ are the same regardless of whether or not the arc is in $\mathbf{A}'_D$; it is easy to verify that for each edge component, the local distributions for these nodes contribute a total of 186 parameters. For each arc $V_i \to V_j \in \mathbf{A}'_D$, the corresponding nodes $C_{ij}$ and $F_{ij}$ contribute a total of $16 + 2 = 18$ parameters; for each arc $V_i \to V_j \notin \mathbf{A}'_D$, the nodes $C_{ij}$ and $F_{ij}$ contribute a total of $4 + 12 = 16$ parameters. For every node $V_i \in \mathbf{V}_D$ in the instance of DBFAS that has exactly two parents, the corresponding $V_i \in \mathbf{X}_L$ in the instance of LEARN will also have two parents. Similarly, for every node $V_i \in \mathbf{V}_D$ with exactly one parent, the corresponding $V_i \in \mathbf{X}_L$ has exactly one parent. By construction of $\mathcal{F}_L$, every parent node for any $V_i \in \mathbf{X}_L$ has two states (and is equal $F_{ji}$ for some $j$), and therefore because each node $V_i \in \mathbf{X}_L$ has nine states, the total number of parameters used in the local distributions for these nodes is $16o_D + 32t_D$. Thus, we conclude that the number of parameters in $\mathcal{F}$ is exactly

$$186|\mathbf{A}_D| + 18|\mathbf{A}'_D| + 16(|\mathbf{A}_D| - |\mathbf{A}'_D|) + 16o_D + 32t_D$$

Because $|\mathbf{A}'_D| \leq k_D$, we conclude from Equation 2 that the number of parameters in $\mathcal{F}_L$ is less than or equal to $d_L$, and thus $\mathcal{F}_L$ is a valid solution to the instance of LEARN. □

**Theorem 9** *There is a solution to the given instance of DBFAS with $|\mathbf{A}'_D| \leq k_D$ if there is a solution $\mathcal{F}_L$ to the instance of LEARN with $\leq d_L$ parameters.*

**Proof:** Given the solution $\mathcal{F}_L$, we create a new solution $\mathcal{F}_L'$ as follows. For every pair $(V_i, V_j)$ corresponding to an edge $V_i \to V_j \in \mathbf{A}_D$ in the instance of DBFAS, if there is no directed path in $\mathcal{F}_L$ from $V_i$ to $V_j$, then the corresponding edge component in $\mathcal{F}_L'$ contains the edges shown in Figure 3a. Otherwise, when there is at least one directed path in $\mathcal{F}_L$ from $V_i$ to $V_j$, the corresponding edge component in $\mathcal{F}_L'$ contains the edges shown in Figure 3b. By construction, $\mathcal{F}_L'$ will contain a cycle only if $\mathcal{F}_L$ contains a cycle, and consequently we conclude that $\mathcal{F}_L'$ is a DAG. From Lemma 7, we know that $\mathcal{F}_L'$ includes $p_L(\mathbf{X}_L)$.

In the next two paragraphs, we argue that $\mathcal{F}_L'$ does not support more parameters than does $\mathcal{F}_L$. Consider the DAG $\mathcal{F}^0$ that is identical to $\mathcal{F}_L'$, except that for all $i$ and $j$, the only parent of $C_{ij}$ is $B_{ij}$ and the only parent of $F_{ij}$ is $E_{ij}$ (see Figure 5). Because $\mathcal{F}^0$ is a subgraph of $\mathcal{F}_L'$, any active path in $\mathcal{F}^0$ must have a corresponding active path in $\mathcal{F}_L'$, and thus we conclude that $\mathcal{F}_L'$ includes $\mathcal{F}^0$. The original solution $\mathcal{F}_L$ also includes $\mathcal{F}^0$ by the following argument: $\mathcal{F}^0$ is a strict sub-graph of $\mathcal{H}_L$ ($\mathcal{F}^0$ contains a subset of the edges and no $H_{ij}$ nodes), and thus any active path in $\mathcal{F}^0$ has a corresponding $\mathbf{X}_L$-restricted active path in $\mathcal{H}_L$; because $\mathcal{H}_L$ is perfect with respect to the distribution $p_L(\mathbf{H}_L, \mathbf{X}_L)$ defined by $(\mathcal{H}_L, \theta_{\mathcal{H}_L})$ (Corollary 6), we know that any such $\mathbf{X}_L$-restricted active path in $\mathcal{H}_L$ corresponds to a dependence in $p_L(\mathbf{X}_L)$, and thus, because $\mathcal{F}_L$ includes $p_L(\mathbf{X}_L)$, there must be a corresponding active path in $\mathcal{F}_L$.

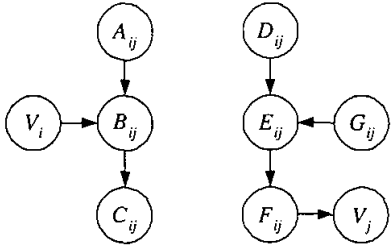

Figure 5: Edges within each edge component of $\mathcal{F}^0$

From Theorem 3, we know that there exists a sequence of edge additions and covered edge reversals that transforms $\mathcal{F}^0$ into $\mathcal{F}_L$, and another sequence of edge additions and covered edge reversals that transforms $\mathcal{F}^0$ into $\mathcal{F}_L'$. From Lemma 1 and Lemma 2, a covered edge reversal does not change the number of parameters supported by a DAG, and thus we can compare the number of parameters supported by the two DAGs by evaluating the increase in parameters that result from the additions within each of the two transformations. $\mathcal{F}^0$ can be transformed into $\mathcal{F}_L'$ by simply adding, for each edge component, the corresponding two extra edges in $\mathcal{F}_L'$. That is, we either (1) add the edges $E_{ij} \to C_{ij}$ and $F_{ij} \to C_{ij}$, resulting in an increase of 12 parameters, or (2) add the edges $B_{ij} \to F_{ij}$ and $C_{ij} \to F_{ij}$, resulting in an increase of 10 parameters. If $\mathcal{F}_L$ supports fewer parameters than $\mathcal{F}_L'$, there must be at least one $\{V_i, V_j\}$ edge component for which the total parameter increase from adding edges between nodes in that component is less than the corresponding increase in $\mathcal{F}_L'$. In order to reverse any edge in an edge component from $\mathcal{F}^0$, we need to first cover that edge by adding at least one other edge that is contained in that component; it is easy to verify that any such "covering addition" results in an increase of at least 16 parameters (adding $E_{ij} \to V_j$ results in this increase, and all other additions result in a larger increase). Thus we conclude that for the $\{V_i, V_j\}$ edge component, only edge additions are performed in the transformation from $\mathcal{F}^0$ to $\mathcal{F}_L$. Because $H_{ij}$ does not exist in $\mathcal{F}_L$, we know that because $p_L(\mathbf{H}_L, \mathbf{X}_L)$ is a DAG-perfect distribution (Corollary 6), $C_{ij}$ and $F_{ij}$ cannot be conditionally independent given any other nodes in $\mathbf{X}_L$; thus, in order for $\mathcal{F}_L$ to include $p_L(\mathbf{X}_L)$, there must be an edge between $C_{ij}$ and $F_{ij}$ in $\mathcal{F}_L$. We consider two cases, corresponding to the two possible directions of the edge between $C_{ij}$ and $F_{ij}$ in $\mathcal{F}_L$. If the edge is directed as $C_{ij} \to F_{ij}$, we know that there is a directed path between $V_i$ and $V_j$ in $\mathcal{F}_L$ because none of the edges from $\mathcal{F}^0$ can be reversed. By construction of $\mathcal{F}_L'$, this implies that the increase in parameters supported by $\mathcal{F}_L'$ attributed to this edge component is 10. The edge $B_{ij} \to F_{ij}$ must exist in $\mathcal{F}_L$, lest there would exist some (possibly empty) set $\mathbf{S} \subset \mathbf{X}_L$ such that $F_{ij} \perp\!\!\!\perp_{\mathcal{F}_L} B_{ij} | C_{ij} \cup \mathbf{S}$ but $F_{ij} \not\!\perp\!\!\!\perp_{p_L} B_{ij} | C_{ij} \cup \mathbf{S}$ (this independence cannot hold in $p_L(\mathbf{X}_L)$ due to the fact that $F_{ij}$ and $B_{ij}$ are d-connected in $\mathcal{H}_L$ given any conditioning set from $\mathbf{X}_L$ that contains $C_{ij}$), contradicting the fact that $\mathcal{F}_L$ includes $p_L(\mathbf{X}_L)$. But adding both $C_{ij} \to F_{ij}$ and $B_{ij} \to F_{ij}$ to $\mathcal{F}^0$ requires an addition of *at least* 10 parameters, contradicting the supposition that the parameter increase due to this edge component is smaller in $\mathcal{F}_L$ than in $\mathcal{F}_L'$. If the edge between $C_{ij}$ and $F_{ij}$ is directed as $F_{ij} \to C_{ij}$, we know that $\mathcal{F}_L$ must also contain the edge $E_{ij} \to C_{ij}$, lest there would exist some conditioning set $\mathbf{S} \subset \mathbf{X}_L$ such that $C_{ij} \perp\!\!\!\perp_{\mathcal{F}_L} E_{ij} | F_{ij} \cup \mathbf{S}$ but $C_{ij} \not\!\perp\!\!\!\perp_{p_L} E_{ij} | F_{ij} \cup \mathbf{S}$, contradicting the fact that $\mathcal{F}_L$ includes $p_L(\mathbf{X}_L)$. Adding both of these edges, however, requires an addition of *at least* 12 parameters; because the corresponding edge component in $\mathcal{F}_L'$ attributed *at most* 12 parameters in the transformation from $\mathcal{F}_L'$, this again contradicts the supposition that the parameter increase due to this edge component is smaller in $\mathcal{F}_L$ than in $\mathcal{F}_L'$.

Having established that $\mathcal{F}_L'$ is a solution to LEARN that supports fewer parameters than $\mathcal{F}_L$, we now use $\mathcal{F}_L'$ to construct a solution $\mathbf{A}_D'$ to the instance of ▶BFAS. For each $\{V_i, V_j\}$ edge component in $\mathcal{F}_L'$, if that component contains the edges shown in Figure 3a, then we include in $\mathbf{A}_D'$ the arc $V_i \to V_j$. $\mathbf{A}_D'$ contains no other arcs.

We now argue that $\mathbf{A}_D'$ contains at least one arc from every cycle from the instance of DBFAS. Each arc $V_i \to V_j \in \mathbf{A}_D$ that is *not* contained in $\mathbf{A}_D'$ has a corresponding edge component in $\mathcal{F}_L'$ for which there is a directed path from $V_i$ to $V_j$. Thus, any hypothetical cycle in the instance of DBFAS that does not pass through an edge in $\mathbf{A}_D'$ has a corresponding directed cycle in $\mathcal{F}_L'$, which is impossible because $\mathcal{F}_L'$ is a DAG.

Finally, we argue that $\mathbf{A}_D'$ contains at most $k_D$ arcs. Recall that $o_D$ and $t_D$ denote the number of nodes in $\mathcal{G}_D$ that have exactly one and two in-coming edges, respectively. As in the proof of Theorem 8, it is easy to verify that the number of parameters $d_L'$ supported by $\mathcal{F}_L'$ is exactly

$$186|\mathbf{A}_D| + 18|\mathbf{A}_D'| + 16(|\mathbf{A}_D| - |\mathbf{A}_D'|) + 16o_D + 32t_D$$

Given that $d_L' \leq d_L$, we conclude from Equation 2 that $|\mathbf{A}_D'| \leq k_D$. □

Given the previous results, the main result of this paper now follows easily.

**Theorem 10** *LEARN is NP-hard.*

**Proof:** Follows immediately from Theorem 8 and Theorem 9. □

### 3.3 EXTENSIONS

In this section, we show that Theorem 10 holds even when the learning algorithm has access to at least one of three oracles, and even when we restrict ourselves to considering only those solutions to LEARN for which each node has at most $k$ parents, for all $k \geq 3$.

The first oracle we consider is an independence oracle. This oracle can evaluate independence queries in constant time.

**Definition 11 (Independence Oracle)**
*An* independence oracle *for a distribution $p(\mathbf{X})$ is an oracle that, in constant time, can determine whether or not $X \perp\!\!\!\perp Y | \mathbf{Z}$ for any $X$ and $Y$ in $\mathbf{X}$ and for any $\mathbf{Z} \subseteq \mathbf{X}$.*

The second oracle we consider can perform certain inference queries in constant time; namely, the inference oracle can return the joint probability of any constant-sized set of variables. This oracle can in turn be used to compute conditional probabilities in constant time using division.

**Definition 12 (Constrained Inference Oracle)**
*A* constrained inference oracle *for a distribution $p(\mathbf{X})$ is an oracle that, in constant time, can compute $p(\mathbf{Z} = \mathbf{z})$ for any $\mathbf{Z} \subseteq \mathbf{X}$ such that $|\mathbf{Z}| \leq k$ for some constant $k$.*

Some learning algorithms use mutual information—or an approximation of mutual information—from a distribution to help construct DAG models. The *(conditional mutual) information* between variables $X$ and $Y$ given the set of variables $\mathbf{Z}$ is defined as

$$Inf(X;Y|\mathbf{Z}) = \sum_{x,y,\mathbf{Z}} p(x,y,\mathbf{z}) \log \frac{p(x,y|\mathbf{z})}{p(x|\mathbf{z})p(y|\mathbf{z})} \quad (4)$$

The third oracle we consider can compute the mutual information between two variable in constant time, given that there are only a constant number of variables in the conditioning set.

**Definition 13 (Constrained Information Oracle)**
*A* constrained information oracle *for a distribution $p(\mathbf{X})$ is an oracle that, in constant time, can compute $Inf(X;Y|\mathbf{Z})$ for any $X$ and $Y$ in $\mathbf{X}$ and for any $\mathbf{Z} \subseteq \mathbf{X}$ such that $|\mathbf{Z}| \leq k$ for some constant $k$.*

**Theorem 14** *Theorem 10 holds even when the learning algorithm has access to (1) an independence oracle, (2) a constrained inference oracle, or (3) a constrained information oracle.*

**Proof:** We establish this result by demonstrating that we can implement all three of these oracles in polynomial time using the parametric Bayesian network $(\mathcal{H}, \theta_\mathcal{H})$ from our reduction. Thus if LEARN can be solved in polynomial time when we have access to any of the constant-time oracles, it must also be solvable in polynomial time *without* any such oracle.

(1) holds immediately because we can test for d-separation in $\mathcal{H}$ in polynomial time. (3) follows from (2) because, given that each variable has some constant number of states, we can implement a constrained information oracle via Equation 4 by calling a constrained inference oracle a constant number of times each.

Let $\mathbf{Z} \subseteq \mathbf{X}$ be any subset of the variables such that $|\mathbf{Z}| \leq k$ for some constant $k$. It remains to be shown how to compute $p(\mathbf{Z} = \mathbf{z})$ in constant time from $(\mathcal{H}, \theta_\mathcal{H})$. The trick is to see that there is always a cut-set of constant size that decomposes $\mathcal{H}$ into a set of polytrees, where each polytree has a constant number of nodes; within any polytree containing a constant number of nodes, we can perform inference in constant time. We define a cut-set $\mathbf{B}$ as follows: $\mathbf{B}$ contains every node $B_{ij}$ for which (1) $C_{ij}$ is in $\mathbf{Z}$ and (2) $B_{ij}$ is not in $\mathbf{Z}$. Note that $\mathbf{B} \cap \mathbf{Z} = \emptyset$. Given conditioning set $\mathbf{B}$, no active path can contain a node $C_{ij}$ as an interior (i.e. non-endpoint) node, even when any subset of $\mathbf{Z}$ is added to the conditioning set (see Figure 6): any such hypothetical active must pass through at least one segment $B_{ij} \rightarrow C_{ij} \leftarrow H_{ij}$. But this is not possible, because every such segment is blocked: if $C_{ij}$ is not in **Z**, then the segment is blocked because $C_{ij}$ has no descendants, and hence can have no descendants in the conditioning set; if $C_{ij}$ is in **Z**, then we know that $B_{ij} \in \mathbf{B}$ and thus the segment is blocked by $B_{ij}$.

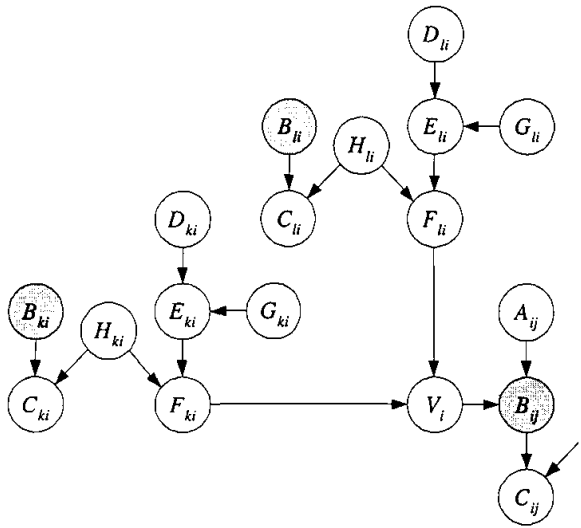

Figure 6: Portion of $\mathcal{H}$ showing that no active path can pass through any $C_{ij}$ once $B_{ij}$ is given.

Because no active path can pass through a node $C_{ij}$, it follows by construction of $\mathcal{H}$ that—given **B** and any subset of **Z**—each node in **Z** is d-connected to only a constant number of other nodes in **Z**. Furthermore, the structure of $\mathcal{H}$ that is bounded between the $C_{ij}$ nodes forms a polytree. Thus, we can express $p(\mathbf{Z} = \mathbf{z})$ as:

$$p(\mathbf{Z} = \mathbf{z}) = \sum_{\mathbf{b}} p(\mathbf{Z} = \mathbf{z}, \mathbf{B} = \mathbf{b})$$
$$= \sum_{\mathbf{b}} \prod_i p(\mathbf{T}_i = \mathbf{t}_i(\mathbf{z}, \mathbf{b}))$$

where each $\mathbf{T}_i$ contains a constant number of variables—$\mathbf{t}_i(\mathbf{z}, \mathbf{b})$ is the set of values for those variables as determined by **z** and **b**—that constitute a polytree in $\mathcal{H}$. Thus, each term $p(\mathbf{T}_i = \mathbf{t}_i(\mathbf{z}, \mathbf{b}))$ above can be computed in constant time using inference in a polytree. Because there are at most $k$ nodes in **Z**, the set **B** can contain at most $k$ nodes. Therefore, given that each node in **B** has at most $r$ states, there are at most $r^k$ terms in the sum above, and we conclude that $p(\mathbf{Z})$ can be computed in polynomial time. □

Finally, we prove that if we restrict LEARN to solutions in which each node has at most $k$ parents, the problem remains NP-hard for all $k \geq 3$.

**Theorem 15** *Theorem 14 holds even when solutions to LEARN are restricted to DAGs in which each node has at most $k$ parents, for all $k \geq 3$.*

**Proof:** The case where $k = 3$ follows immediately from the proof of Theorem 8, where the constructed solution to LEARN is a DAG in which each node has at most three parents, and from the proof of Theorem 9, where the given solution to LEARN is converted into a (better) solution in which each node has at most three parents. It is easy to see that these proofs remain valid under a less restrictive ($k > 3$) bound on the number of parents, and thus the theorem follows. □

## 4 CONCLUSION

In this paper, we demonstrated that the problem of identifying high-scoring DAG models from large datasets when using a consistent scoring criterion is NP-hard. Together with the result of Chickering (1996) that the non-asymptotic learning problem is NP-hard, our result implies that learning is hard regardless of the size of the data. There is an interesting gap in the present results. In particular, Chickering (1996) proved that finite-sample learning is NP-hard when each node is restricted to have at most two parents, whereas in this paper we proved that large-sample learning is NP-hard with a three-parent restriction. This leads to the question of whether or not large-sample learning is NP-hard when we restrict to two parents; we believe that this problem is probably NP-hard, and is worth further investigation.

In practice, the large-sample learning problem actually requires scanning a dataset with a large number of samples, as opposed to accessing a compact representation of the generative distribution. We could alternatively have defined a learning problem in which there is an actual data set supplied; the problem with this approach is that in order to guarantee that we get the large-sample ranking of models, we will need the number of data points to be so large that the size of the problem instance is exponential in the number of variables in the domain. Our results have practical importance when it is reasonable to assume that (1) there is enough data such that the relative ranking of those DAG models considered by the learning algorithm is the same as in the large-sample limit, and (2) the number of records in the data is small enough that we can compute the score for candidate models in a reasonable amount of time.

As discussed in Section 1, there exist assumptions about the generative distribution that lead to efficient large-sample learning algorithms. These assumptions are not likely to hold in most real-world scenarios, but the corresponding "correct" algorithms can work well even if the assumptions do not hold. An interesting line of research is to investigate alternative, weaker assumptions about the generative distribution that lead to efficient learning algorithms and guarantee large-sample correctness.

## Acknowledgments

We thank Dimitris Achlioptas and Ramarathnam Venkatesan for useful discussions.

## References

[Bouckaert, 1994] Bouckaert, R. R. (1994). Properties of Bayesian belief network learning algorithms. In *Proceedings of Tenth Conference on Uncertainty in Artificial Intelligence,* Seattle, WA, pages 102–109. Morgan Kaufmann.

[Chickering, 1995] Chickering, D. M. (1995). A transformational characterization of Bayesian network structures. In Hanks, S. and Besnard, P., editors, *Proceedings of the Eleventh Conference on Uncertainty in Artificial Intelligence,* Montreal, QU, pages 87–98. Morgan Kaufmann.

[Chickering, 1996] Chickering, D. M. (1996). Learning Bayesian networks is NP-Complete. In Fisher, D. and Lenz, H., editors, *Learning from Data: Artificial Intelligence and Statistics V,* pages 121–130. Springer-Verlag.

[Chickering, 2002] Chickering, D. M. (2002). Optimal structure identification with greedy search. *Journal of Machine Learning Research*, 3:507–554.

[Chickering et al., 2003] Chickering, D. M., Meek, C., and Heckerman, D. (2003). Large-sample learning of Bayesian networks is NP-hard. Technical Report MSR-TR-2003-17, Microsoft Research.

[Dasgupta, 1999] Dasgupta, S. (1999). Learning polytrees. In Laskey, K. and Prade, H., editors, *Proceedings of the Fifteenth Conference on Uncertainty in Artificial Intelligence,* Stockholm, Sweden, pages 131–141. Morgan Kaufmann.

[Garey and Johnson, 1979] Garey, M. and Johnson, D. (1979). *Computers and intractability: A guide to the theory of NP-completeness.* W.H. Freeman.

[Garvil, 1977] Garvil, F. (1977). Some NP-complete problems on graphs. In *Proc. 11th Conf. on Information Sciences and Systems,* Johns Hopkins University, pages 91–95. Baltimore, MD.

[Meek, 1997] Meek, C. (1997). *Graphical Models: Selecting causal and statistical models.* PhD thesis, Carnegie Mellon University.

[Meek, 2001] Meek, C. (2001). Finding a path is harder than finding a tree. *Journal of Artificial Intelligence Research*, 15:383–389.

[Pearl, 1988] Pearl, J. (1988). *Probabilistic Reasoning in Intelligent Systems: Networks of Plausible Inference.* Morgan Kaufmann, San Mateo, CA.

[Spirtes et al., 2000] Spirtes, P., Glymour, C., and Scheines, R. (2000). *Causation, Prediction, and Search (second edition).* The MIT Press, Cambridge, Massachussets.